\documentclass{article} %
\usepackage{arxiv}

\usepackage{microtype}
\usepackage{hyperref}
\usepackage{url}
\usepackage{amsmath}
\usepackage{amsfonts}
\usepackage{booktabs}
\usepackage{cleveref}
\usepackage{graphicx}
\graphicspath{{./}{../}}
\usepackage{amssymb}
\usepackage{algorithm}
\usepackage[noend]{algorithmic}
\usepackage{wrapfig}
\usepackage[normalem]{ulem}
\usepackage{pifont}

\usepackage{tcolorbox, listings, bm}
\tcbuselibrary{skins, breakable, listings, raster}

\usepackage{lineno}
\usepackage{xspace}
\definecolor{scggreen}{HTML}{DCEEF0}
\definecolor{scgborder}{HTML}{458588}
\definecolor{norevred}{HTML}{E4F0E4}
\definecolor{norevborder}{HTML}{689D6A}
\definecolor{seedgray}{HTML}{F1F3F5}
\definecolor{seedborder}{HTML}{6C757D}
\definecolor{orhighlight}{HTML}{C62828}
\definecolor{codeblue}{HTML}{1565C0}
\definecolor{codegray}{HTML}{455A64}
\definecolor{conclusionhl}{HTML}{1565C0}
\definecolor{ours}{HTML}{51969A}
\definecolor{RL}{HTML}{D65E0D}
\definecolor{solved}{HTML}{98971A}
\definecolor{notsolved}{HTML}{cc241d}

\definecolor{darkblue}{rgb}{0, 0, 0.5}
\hypersetup{colorlinks=true, citecolor=darkblue, linkcolor=darkblue, urlcolor=darkblue}

\newcommand{\algofullname}{Self-Guided Self-Play}
\newcommand{\algoname}{SGS\xspace}
\newcommand{\baseline}{$\text{REINFORCE}^{1/2}$\xspace}
\newcommand{\reviewer}{Guide\xspace}
\newcommand{\generator}{Conjecturer\xspace}
\newcommand{\solver}{Solver\xspace}

\title{Scaling Self-Play with Self-Guidance}

\author{Luke Bailey, Kaiyue Wen, Kefan Dong, Tatsunori Hashimoto, Tengyu Ma 
}

\usepackage{xcolor}
\definecolor{infra-blue}{HTML}{0076BA}
\definecolor{infra-fault}{HTML}{F27300}

\begin{document}

\maketitle

\begin{center}
    \vspace{-2mm}
   Stanford University  
\end{center}

\newcommand{\tnote}[1]{{\color{blue} TM: #1}}
\begin{abstract}
    LLM self-play algorithms are notable in that,
    in principle, nothing bounds their learning: a \generator model creates
    problems for a \solver, and both improve together.
    However, in practice, existing
    LLM self-play methods \emph{do not scale well 
    with large amounts of compute}, instead hitting learning plateaus.
    We argue this is because over long training runs,
    the \generator learns to hack its reward, 
    collapsing to artificially
    complex problems that do not help the \solver improve. To overcome this, we introduce
    \algofullname{} (\algoname), a self-play algorithm in which the
    language model itself guides the \generator away from degeneracy. In \algoname, the
    model takes on three roles: \solver, \generator, and a \emph{\reviewer}
    that scores synthetic problems by their relevance to unsolved target problems
     and how clean and natural 
    they are,
    providing supervision against \generator collapse.
    Our core
    hypothesis is that language models can assess whether a subproblem is
    useful for achieving a goal. We evaluate the scaling properties 
    of \algoname by running training 
    for significantly
    longer than prior works and  by fitting
    scaling laws to cumulative solve rate curves. 
    Applying
    \algoname to formal theorem proving in Lean4, we find that it
    surpasses the asymptotic 
    solve rate of our strongest 
    RL baseline in fewer
    than 80 rounds of self-play
    and enables a 7B
    parameter model, after 200 rounds of self-play, to solve more problems 
    than a 671B parameter model pass@4.
\end{abstract}

{
\renewcommand{\thefootnote}{}
\footnotetext{Correspondence to \href{mailto:ljbailey@stanford.edu}{ljbailey@stanford.edu}\quad\quad\quad
Code available at \url{https://github.com/LukeBailey181/sgs}}
\addtocounter{footnote}{-1}
}

\begin{figure}[h]
    \centering
    \includegraphics[width=0.95\textwidth]{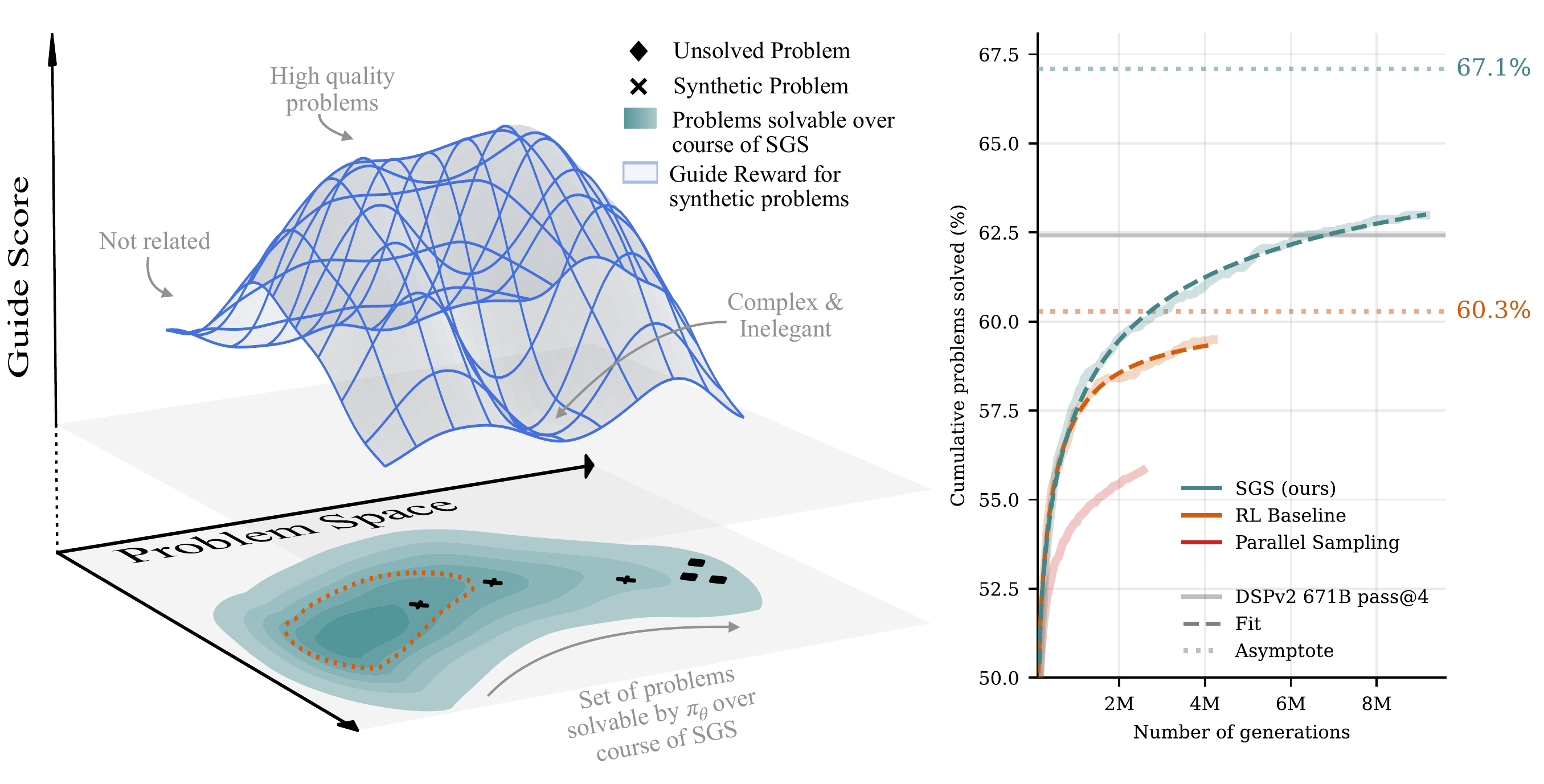}
    \caption{
        \textbf{Left:} Intuition behind \algoname. 
        Bottom, the space of problems solvable over course of \algoname, 
        starting as a \textbf{\textcolor{ours}{small area}}, expanding past 
        \textbf{\textcolor{RL}{\dotuline{asymptotic RL performance}}}, 
        directed towards unsolved target problems ($\blacklozenge$) by synthetic problems ($\bm{\times}$). 
        Top, the \reviewer score for synthetic problems, which is 
        high along the path to unsolved targets, and low for not related 
        and inelegant problems (see \Cref{fig:qualitative}).
        \textbf{Right:}
        Results of running \algoname on $D_{\text{3k}}$ ($\approx 3000$ Lean4
        formal math problems) compared to
        our best performing RL 
        baseline 
        (\Cref{sec:algoname_performance}). 
        At 6.3M generations, \algoname applied to the 7B parameter 
        DeepSeek-Prover-V2 model exceeds the pass@4 of the larger 671B counterpart (DSPv2 671B).
        }
    \label{fig:figure_1}
\end{figure}

\section{Introduction}

We would like AI systems that, given enough compute, 
can solve any problem 
we give them.
Unfortunately, when a problem
is sufficiently difficult,
standard reinforcement learning techniques fail 
due to sparse or absent reward \citep{prakash2025can,qu2026pope}. 
Asymmetric self-play offers a solution
to the sparse reward problem. In this framework, a \generator model
proposes new learning problems
that bridge the gap between the current \solver capability and the
unsolved problems
\citep{sukhbaatar2017intrinsic,florensa2018automatic}.
Over time, the \solver and \generator
improve together. 

Self-play algorithms are unique in that,
in principle, nothing bounds their learning \citep{silver2017mastering}.
However, existing LLM self-play methods 
do not sustain 
learning over long periods 
\citep{yu2025guided,chae2025towards}. In general, these 
methods reward the synthetic problems only
according to the \solver{}'s pass rate on the problem
\citep{zhao2505absolute,huang2025r,chen2025self,liu2025spice,yang2026ttcs,jana2026gasp,kuba2025language}.
Over
time the \generator hacks this reward function, collapsing 
to artificially complex problems that are
not useful for learning (\Cref{fig:qualitative}) \citep{dong2025stp}. 
That is,  \emph{the quality of the synthetic problems degrades
over the course of training}, leading to a learning plateau.

We can quantify this plateauing problem by fixing a set of hard training problems
and seeing how many we can solve with self-play. 
We strive for an algorithm that, if you run it for long enough, 
leads to all the problems being solved. In the pursuit of this 
ideal, we ask the question: 

\begin{center}
\emph{
    \textbf{Given a fixed set of hard problems, 
    how can we scale self-play\\to solve as many as possible?}
}
\end{center}

Towards this end, we propose \algofullname{} (\algoname). 
\algoname is an asymmetric self-play algorithm in which an LLM 
takes on three roles, \solver, \generator, and \reviewer. 
It has two mechanisms to stop \generator degradation over the course of training.
First, for each unsolved problem, the \generator is prompted 
to produce a synthetic problem that is  
useful for solving that problem. Secondly, the \reviewer is used to score each 
synthetic problem by how relevant it is to its corresponding unsolved problem, 
and how clearly it is 
formulated.
Crucially, a synthetic problem that is superficially related to a target but is messy, overly complex, or inelegant, is downweighted by the \reviewer. 
The \solver is rewarded for
every problem it solves, while the \generator
is rewarded for producing problems that the \solver can make progress on
and that the \reviewer judges to be high quality. 
Our core hypothesis 
is that language models can assess whether a subproblem is 
useful for achieving a goal, and thus guide the learning 
process towards that end.

To evaluate the scaling dynamics of \algoname, we take two complementary
approaches. First, we run training for much longer than prior works: our main experiment
produces over 6 billion tokens during the self-play procedure,
epoching the target data over 230 times
\citep{zhao2505absolute,huang2025r,chen2025self}. 
Second, we fit scaling laws to the cumulative solve rate over
training, allowing us to extrapolate long-run behavior and compare the asymptotic
solve rates of different methods \citep{khatri2025art,ruan2024observational}.

We use this evaluation method to analyze the long-running dynamics of \algoname
applied to formal theorem proving in Lean4, where the 
goal is to generate formally verified proofs of mathematical statements
\citep{de2015lean,moura2021lean}. 
In this setting, the target problems are Lean theorems, and solutions are proofs 
that can be automatically checked by the Lean compiler. We provide a summary 
of our contributions below:\\

\begin{itemize}
    \item In \Cref{sec:algoname_performance}, we show that \algoname
    outperforms RL  and parallel sampling baselines, achieving a
    7\% higher asymptotic solve rate than RL alone. With sufficient compute,
    \algoname enables DeepSeek-Prover-V2-7B to surpass the pass@4 of the
    much larger DeepSeek-Prover-V2-671B model \citep{ren2025deepseek}.
    \item 
    In \Cref{sec:reviewer_study}, we study the design of the \generator
    training algorithm. We ablate conditioning the \generator 
    on unsolved problems and the \reviewer component. In both 
    cases we find the \generator collapses to producing 
    degenerate problems that do not help the \solver make progress 
    on the training problems. 
    We also show that freezing the
    \generator is suboptimal, as the \solver quickly saturates the fixed
    problem distribution.
    \item In \Cref{sec:entropy_collapse}, we study the design of the \solver
    training algorithm. We show that \solver entropy collapse can starve the
    \generator of training signal, making the choice of \solver RL objective
    critical for long-running self-play.
\end{itemize}

Overall, our key findings are as follows: (1) scaling asymmetric self-play 
requires avoiding undesirable collapses in the \solver and \generator distributions,
which can be achieved by (2) selecting the correct \solver objective, 
conditioning the \generator on unsolved problems, and adding a data quality 
reward. We instantiate 
these ideas in \algoname, demonstrating that LLMs themselves 
can be used as the data quality metric.

\section{Related Work}

\textbf{Asymmetric Self-Play.}
\algoname is an instance of
\emph{asymmetric self-play},
methods
in which agents with asymmetric roles, typically 
a \generator generating tasks and \solver solving them, learn through interaction.
This paradigm traces its roots to 
\citet{sukhbaatar2017intrinsic} and
\citet{florensa2018automatic},
who demonstrate that a proposer can generate goals 
for a solver in navigation and control environments.
Asymmetric self-play hails from earlier work 
on self-play, involving an agent playing
a zero-sum game against itself \citep{samuel1959some,tesauro1995temporal,thrun1994learning,silver2016mastering}.

More recently, the 
asymmetric self-play paradigm has 
been applied to tasks 
with natural language action 
spaces using LLMs. Some use 
real data to ground the generated 
problems \citep{liu2025spice,yu2025guided,choianchored,sundaram2026teaching,jana2026gasp,yang2026ttcs}, and others
 consider \generator{}s that
use \emph{no seed data} 
\citep{huang2025r,zhao2505absolute,xia2025agent0,chen2025self,kuba2025language,li2026r,dineen2026vocabulary}. 
In terms of application domain, 
\citet{dong2025stp} and \citet{poesia2024learning}
are most closely related to our work. Both 
apply asymmetric self-play to formal theorem 
proving. \citet{dong2025stp} generate formal 
statements conditioned on \emph{solved problems}.
\citet{poesia2024learning} consider a 
setting where an agent builds up mathematics from scratch starting with only axioms.

Our work differs from  prior works  
in three ways. Firstly, the majority of existing 
algorithms \emph{only reward the \generator according
to the \solver performance}, which we find leads to
synthetic problem degradation (\Cref{sec:reviewer_study}, 
\Cref{fig:qualitative}). We address this using the \reviewer model. 
We only see this degradation issue because secondly, we scale the self-play 
procedure to longer runs than prior works. We do 
so because thirdly, we are motivated to study 
the scaling properties of self-play to solve problems
far beyond the base model capability, and thus overcome 
learning plateaus.

\textbf{LLMs for Formal Math.}
We use formal math as the application 
area to test \algoname. 
The current state-of-the-art methods that apply 
LLMs to formal mathematics use complex inference 
time strategies and agent scaffolds
\citep{chen2025seed1, chen2025seed2, varambally2025hilbert}. 
In this work, we use the simpler whole proof
generation method, where an LLM takes as input 
the formal problem, and generates an entire proof directly,
possibly after some natural language reasoning 
\citep{lin2025goedel1, lin2025goedel2, ren2025deepseek}.
There is no reason, however, that \algoname could not be used
to train a complex agent-based prover, but we leave this to future work.

\section{The \algoname Algorithm}
\label{sec:method}

We consider a fixed set of problems
$\mathcal{D} = \{x_1, \dots, x_N\}$
which we aim to solve.  
We assume each problem admits 
a verifiable solution; in our work,
problems are formal math questions in Lean4. Our goal is to solve
as many of 
the problems in $\mathcal{D}$ as possible given a large compute budget.
\algoname has three components: (1) a \solver $\pi_\theta$,
(2) a Synthetic Data \generator $g_\phi$, and (3) a
\reviewer $\rho$, an LLM-based judge that evaluates
synthetic data quality. All of these are \emph{initialized from
the same base model}. Each iteration of the algorithm
proceeds as follows (see Algorithm~\ref{alg:cerberus}):

\begin{enumerate}
    \item \textbf{Synthetic problem generation}: Sample a batch $\mathcal{B} \subseteq \mathcal{D}$ of problems. For each problem $x \in B$ not yet solved by the \solver, the \generator produces a related synthetic problem $\tilde{x} \sim g_\phi(\cdot \mid x)$, 
    which is intended to be simpler but meaningfully related. In full generality, the
    \generator would have to produce an entire Markov
    Decision Process (MDP), including a reward function,
    to train the \solver. In our work, the formal
    problem produced by the \generator comes 
    with a reward function, the Lean4 compiler.
    \item \textbf{Rollout and verification}: The \solver attempts to solve both the original
    problems $x \in \mathcal{B}$ and the synthetic problems $\tilde{x}$, and rewards are calculated.
    \item \textbf{\solver and \generator update}: the \solver is updated using a reinforcement learning objective over rollouts on both original and synthetic problems.
    The \generator is updated using a reward that favors useful synthetic problems;
    for each $\tilde{x}$, the reward combines (a) a difficulty signal proportional to the
    \solver{}’s empirical solve rate on $\tilde{x}$, favoring problems of intermediate
    difficulty, and (b) a score $\rho(x, \tilde{x}) \in [0,1]$ produced by
    the \reviewer LLM, indicating how well $\tilde{x}$ relates to its target problem $x$, and how 
    elegant a conjecture it is.
\end{enumerate}

\subsection{Synthetic Data Generation and Verification (Steps 1, 2)}
\label{sec:data_generation}

We partition the sampled batch of data 
$\mathcal{B} = \{x_i\}_{i=1}^B \subseteq \mathcal{D}$ into
two subsets: $\mathcal{B}_{\mathrm{solved}}$, containing problems 
with a correct solution from prior rounds,
and $\mathcal{B}_{\mathrm{unsolved}}$, containing unsolved problems. 
For each problem $x \in \mathcal{B}_{\mathrm{unsolved}}$, the
\generator $g_\phi$ produces a synthetic problem
$\tilde{x} \sim g_\phi(\cdot \mid x)$. The \generator
is prompted to create a version of $x$ that is conceptually
related but simpler to solve  (see Appendix \Cref{sec:appendix_prompting_details}
for prompt).
For problems in $\mathcal{B}_{\mathrm{solved}}$, no synthetic problems are generated. Let $\mathcal{B}_\mathrm{synth}$ 
be the set of synthetic problems generated.
For each problem in $\mathcal{B} \cup \mathcal{B}_\mathrm{synth}$, the \solver
$\pi_\theta$ generates $k$
attempts. 
For each solution $y_x^i$ for
problem $x$, we determine its correctness $v(y_x^i) \in \{0,1\}$
(for our experiments, using the Lean4 compiler). 

\subsection{\solver and \generator Update (Step 3)}
Next we update the
parameters of the \solver $\pi_\theta$ and the \generator
$g_\phi$ using the collected trajectories. For
our experiments, we do not tie the weights of
the \solver and \generator and thus we perform
separate optimization steps.\footnote{Tying 
weights between \solver, \generator, and \reviewer is 
entirely possible, however we do not explore it in this work.}

\begin{wrapfigure}{r}{0.58\textwidth}
    \vspace{-10pt}
    \begin{minipage}{0.58\textwidth}
    \begin{algorithm}[H]
    \caption{\algoname}
    \label{alg:cerberus}
    \begin{algorithmic}[1]
    \REQUIRE Dataset $\mathcal{D}$, \solver $\pi_\theta$, \generator $g_\phi$, \reviewer $\rho$
    \STATE Initialize all problems in $\mathcal{D}$ as unsolved
    \FOR{iteration $t = 1, 2, \ldots$}
        \STATE Sample batch $\mathcal{B} \subseteq \mathcal{D}$
        \STATE Split $\mathcal{B}$ into $\mathcal{B}_{\mathrm{solved}}, \mathcal{B}_{\mathrm{unsolved}}$
        \FOR{each $x \in \mathcal{B}_{\mathrm{unsolved}}$}
            \STATE Generate $\tilde{x} \sim g_\phi(\cdot \mid x)$
        \ENDFOR
        \STATE Collect synthetic set $\mathcal{B}_{\mathrm{synth}}$
        \FOR{each problem $x \in \mathcal{B} \cup \mathcal{B}_{\mathrm{synth}}$}
            \STATE Sample $k$ solutions from $\pi_\theta$
            \STATE Verify each solution $v(y_x^i) \in \{0,1\}$
        \ENDFOR
        \FOR{each $\tilde{x} \in \mathcal{B}_{\mathrm{synth}}$}
            \STATE $s(\tilde{x}) \gets  \text{solve rate of } \tilde{x} $
            \STATE $\text{ind} \gets \mathbf{1}[s(\tilde{x}) \neq 0 \land s(\tilde{x}) \text{ in bottom }70\%]$
            \STATE $R_{\text{solve}} \gets  \text{ind} \cdot  (1 -s (\tilde{x}))$
            \STATE $R_{\text{guide}} \gets \rho(x, \tilde{x})$
        \ENDFOR
        \STATE Update \solver $\pi_\theta$ with RL using $v(y_x^i)$
        \STATE Update \generator $g_\phi$ with RL using $R_{\text{solve}} \cdot R_{\text{guide}}$
    \ENDFOR
    \end{algorithmic}
    \end{algorithm}
    \end{minipage}
    \vspace{-4mm}
\end{wrapfigure}

\textbf{\solver Update.}
The \solver $\pi_\theta$ is updated on correct solutions. For each rollout $y_x^i$ the
reward is its binary verification $v(y_x^i)$.
In principle, any reinforcement learning objective can be 
used to update the \solver using $v(y_x^i)$.
We use a REINFORCE objective on all problems
with solve rate less than or equal to 0.5, promoting 
learning on hard problems. As 
the reward is binary, this reduces to a 
simple log likelihood objective over all correct rollouts 
for problems with solve rate less than or equal to 0.5
(see \Cref{sec:appendix_rl_objective_functions} for more details).
We refer to this objective as \baseline,
and 
justify its use over other RL objectives in \Cref{sec:algoname_performance}.

\textbf{\generator Update.}
The goal of the \generator is to produce synthetic problems $\tilde{x}$ that are
both relevant to the conditioned target $x$ and neither too simple nor 
impossible. We define the reward for a synthetic
problem as $R_{\text{synth}} = R_{\text{solve}} \cdot R_{\text{guide}}$.
To update the conjecturer with $R_\mathrm{synth}$ we linearly normalize
it within its batch to [0,1] (see \Cref{sec:app_conj_objective}).

\emph{Solve Rate Reward ($R_{\text{solve}}$):} Let
$s(\tilde{x}) = \frac{1}{k} \sum_{i=1}^k
v(y^i_{\tilde{x}})$ be the solve rate
of the synthetic problem. We set $R_{\text{solve}} = 0$
if $s(\tilde{x}) = 0$ (too difficult) or if $s(\tilde{x})$
is in the top 30\% of solve rates within the current batch
(too easy). For all other problems, we set $R_{\text{solve}} 
= 1 - s(\tilde{x})$, thereby favoring harder problems within
the solvable range.

\emph{\reviewer Reward ($R_{\text{guide}}$):} 
We use a \reviewer model $\rho$ that is 
the same model as the initial \solver and \generator.
The \reviewer is prompted to evaluate the quality
of $\tilde{x} \sim g_\phi(\cdot|x)$ relative to the unsolved $x$,
outputting a score $R_{\text{guide}} = \rho(x,\tilde{x})$. 
The details 
of this prompt greatly affect the 
synthetic data. For our experiments, we ran 
a long testing run, iteratively viewing generated 
problems, identifying flaws in them, and prompting the \reviewer 
to down-weight such problems. In the end, we 
had a fairly simple rubric that outputted 
high scores for problems that were (1) related to 
the unsolved problem, and (2) were
formulated clearly by having a simple 
conclusion and 
no redundant premises. We provide further details, 
and the \reviewer prompts used, in \Cref{sec:appendix_prompting_details}.
Like the \solver, we use a simple REINFORCE objective to
update the \generator. We provide details in 
Appendix \Cref{sec:appendix_rl_objective_functions}.

\section{Experimental Results}
\label{sec:ex}

\begin{figure}[t]
    \centering
    \includegraphics[width=1\textwidth]{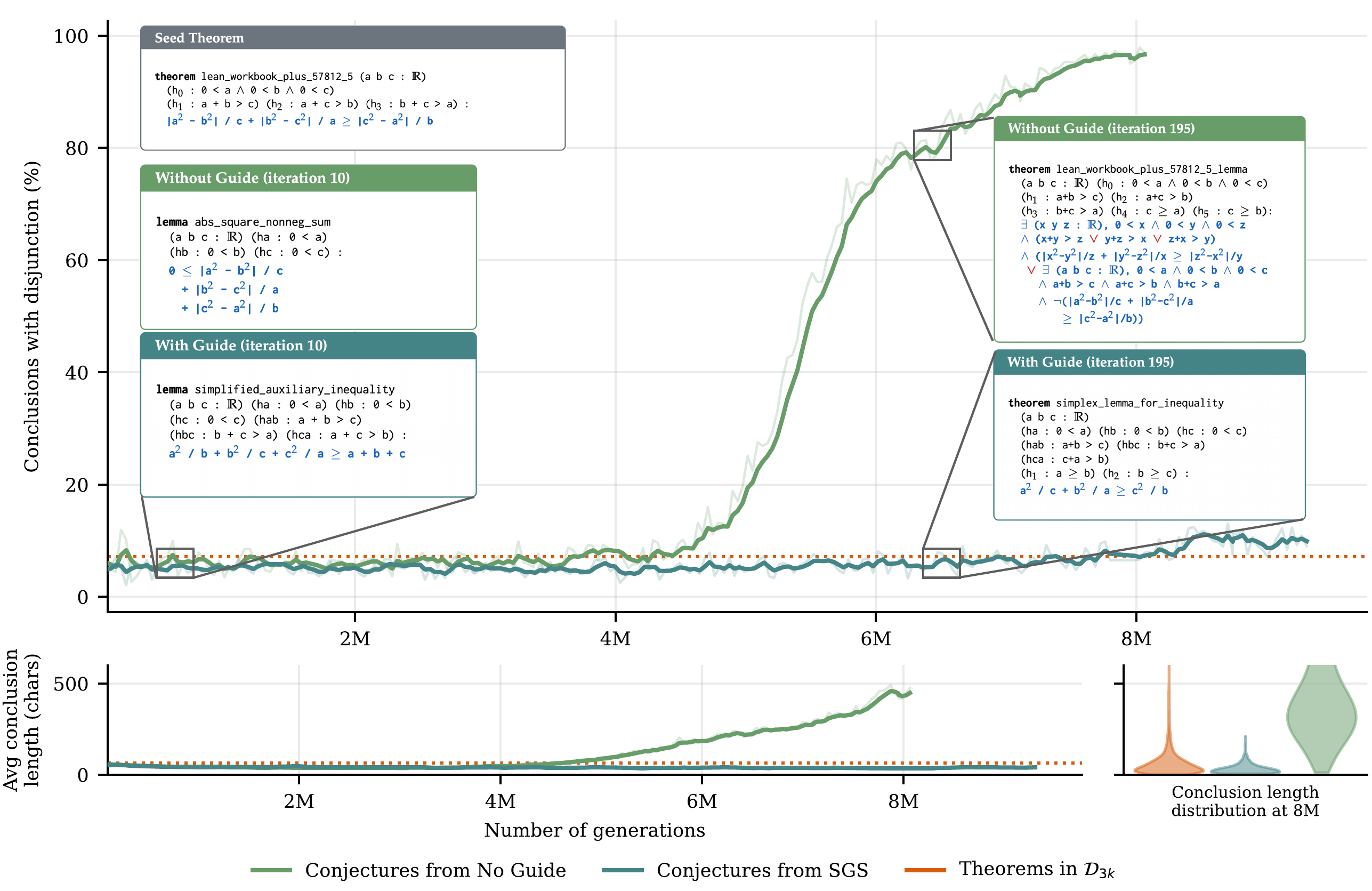}
    \caption{
        \textbf{Benefit of \reviewer component in \algoname}.
        \textbf{Top:} Percentage of generated problems with disjunctive conclusions.
        Without the \reviewer, this rises to over 80\%; \algoname stays near the
        $D_{\text{3k}}$ baseline (\textbf{\textcolor{RL}{\dotuline{dotted orange}}}).
        \textbf{Bottom:} Average conclusion length,
        confirming that No \reviewer problems become overly long and complex.
        \textbf{Inlaid examples:} Problems generated from the same 
        unsolved seed theorem, problem conclusions 
        are shown in \textbf{\textcolor{conclusionhl}{blue}}. 
        By iteration~195, \algoname generates a related problem that assumes an ordering
        of side lengths (such an assumption is required to solve the seed problem with cases) while No \reviewer
        produces a convoluted statement with disjunctions
        (\textcolor{orhighlight}{$\bm{\lor}$}).
    }
    \label{fig:qualitative}
\end{figure}

\subsection{Implementation Details}
\label{sec:implementation}

\textbf{Dataset and training details.} For our main experiments, we use a subset 
of the Goedel-Pset-V1 \citep{lin2025goedel1}, a dataset of auto formalized 
problems primarily from the NuminaMath-CoT dataset \citep{li2024numinamath} 
(itself a collection of math problems spanning from grade school to undergraduate level). 
Our experiments are focused on trying to solve a small fixed set of problems 
over a long training run. Accordingly, we sample 5,000 problems from Goedel-Pset-V1.
Auto formalization methods can result in problems that are impossible
to prove due to a formalization error. Because of this we remove 
any problems that GPT 5 mini determines are
impossible to prove due to a formalization error. This results in 
our main experimental dataset of 3,323 problems, hereafter 
referred to as $D_{\text{3k}}$ (for more details see 
Appendix \Cref{sec:appendix_data_details}).
For all experiments, we set our batch size to the total number of 
problems (thus we sample 
rollouts for all 3,323 problems), and do 8 proof rollouts per problem. We use the full batch setting because we 
get high hardware utilization by running a longer rollout 
phase, and we remove any batch variance from metrics. 
We use custom training infrastructure to utilize a heterogeneous 
GPU and CPU cluster, detailed in Appendix \Cref{sec:appendix_training_infrastructure}.

We apply a length penalty to all RL updates 
inspired by Soft Overlong Punishment \citep{yu2025dapo}.
Rollouts that consume between 80\% and
100\% of the context window receive a negative reward that scales
linearly from 0 to $-1$. We also give 0 reward to any \solver proof with the \texttt{try} tactic, 
as we found it common for the \solver to get caught in infinite loops using this tactic.

\textbf{Model.} We use 
DeepSeek-Prover-V2-7B \citep{ren2025deepseek}, 
a model trained with 
large-scale RL for Lean4, to initialize the \solver, \generator, and \reviewer in \algoname.
Because of the Lean-specific RL, the model has poor instruction
following capabilities.
In the \generator role, at the start of \algoname
it outputs the correct format to extract a synthetic problem 51.5\% of the time,
which is acceptable as the \generator RL 
increases this. For the \reviewer, the base model 
outputs the correct format 54.7\% of the time. As the \reviewer is not trained, this 
error rate would persist, and is thus unacceptable. To alleviate this,
we run a small SFT stage. We take 2048 examples of well formatted
\reviewer outputs, created using GPT 4.1 mini, and finetune DeepSeek-Prover-V2-7B on them. 
This increases the number of well formatted \reviewer outputs to over 99\%. 

\begin{figure}[t]
    \centering
    \includegraphics[width=1\textwidth]{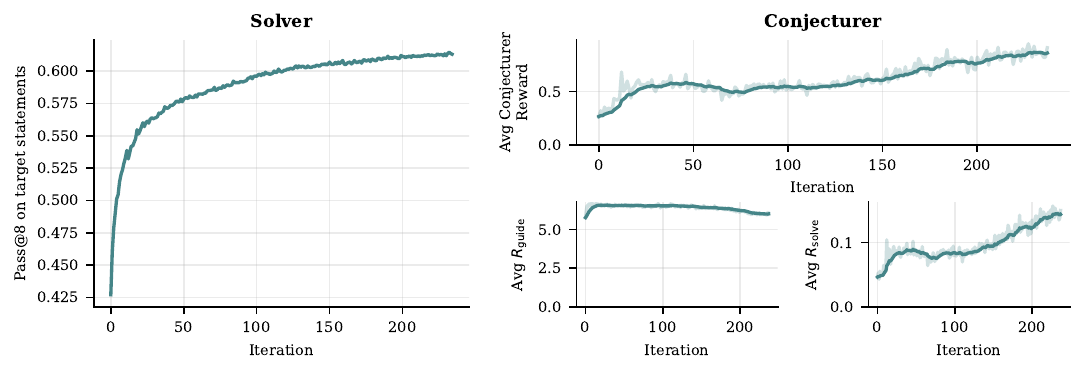}
    \caption{
        Training dynamics of \algoname. \textbf{Left:} \solver pass@8 on target
        statements increases steadily. \textbf{Right:} \generator metrics, average \generator reward (top,
    varies between 0 and 7), average Guide reward (bottom left, varies between 0 and 8), and average solve rate reward
    (bottom right, varies between 0 and 7/8).
    We see $R_{\text{guide}}$ begins high (due to initial \generator 
    prompt providing a good prior)
    and \emph{critically
    remains there}, while $R_{\text{solve}}$ 
    increases during training.
    }
    \label{fig:main_run_reward}
\end{figure}

\subsection{Modeling scaling performance}
\label{sec:modeling_scaling_performance}

Unlike pre-training scaling laws which model loss, our 
setting requires modeling a bounded accuracy metric, specifically 
the cumulative solve rate on training problems. Following other 
works modeling accuracy metrics \citep{khatri2025art, ruan2024observational}, we adopt a 
sigmoidal curve (with respect to log compute) of the form

\begin{align}
    R_C = R_0 + (A - R_0) \cdot \frac{1}{1 + (C_{\text{mid}} / C)^B},
\end{align}

where $R_C$ is the cumulative solve rate after
$C$ model generations (our proxy for compute),
$R_0$ is the initial solve rate, 
$A$ is
the asymptotic cumulative solve rate, $C_{\text{mid}}$ is the curve midpoint,
and $B$ controls the rate at which performance approaches the asymptote. 
We provide details of how we fit the scaling laws in Appendix \Cref{sec:appendix_fitting_cumulative_solve_rate}.

\textbf{Validation of fits}. Our fits are stable when 
removing 10\%, 20\%, and 30\% of the final training data, and are
robust when randomly dropping 50\% of training points: the fitted asymptote varies 
by 1.1\% (\Cref{fig:fit_sensitivity}, Appendix \Cref{sec:appendix_fitting_cumulative_solve_rate}). 
We therefore treat differences 
below 1.1\% cautiously. Since true asymptotic solve rates are unobservable, 
we use fitted scaling laws as a useful \emph{additional} metric rather than a 
substitute for long-run empirical performance.

\subsection{\algoname performance}
\label{sec:algoname_performance}

\begin{wrapfigure}{r}{0.5\textwidth}
    \vspace{-5mm}
    \centering
    \includegraphics[width=0.5\textwidth]{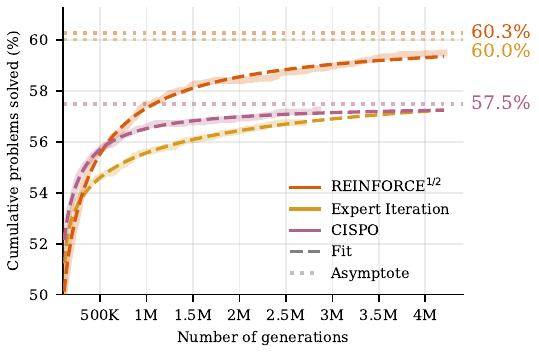}
    \vspace{-6mm}
    \caption{
        Performance of each RL baseline method on $D_{\text{3k}}$.
        \baseline performs best, closely followed by EI.
        CISPO performs worst due to entropy collapse, shown
        in \Cref{fig:baselines_entropy}.
    }
    \label{fig:baselines}
    \vspace{-5mm}
\end{wrapfigure}

\textbf{RL baselines.}
We begin by carefully benchmarking three different RL baselines on $D_{\text{3k}}$. 
As a representative example 
of grouped RL objectives, the  dominant paradigm of LLM RL, 
we use CISPO \citep{chen2025minimax}. We select this over other 
grouped RL objectives as \citet{khatri2025art} demonstrate 
it has better asymptotic validation accuracy than other methods.
We also test a variant of Expert Iteration suggested by \citet{dong2025stp}, 
which involves only sampling solutions for any problem that we have solved fewer 
than 16 times. Finally, we test 
\baseline as introduced in \Cref{sec:method}. For 
each algorithm, we tune the learning rate on $D_{\text{3k}}$ 
to ensure optimal performance. We provide 
 further details in \Cref{sec:appendix_rl_objective_functions}.
Overall we find \baseline is the best performing (\Cref{fig:baselines}).
CISPO performs poorly due to entropy collapse (\Cref{fig:baselines_entropy}).
In \cref{sec:entropy_collapse} we show that this entropy collapse 
makes vanilla CISPO unsuitable as a \solver update objective for \algoname.

\begin{figure}
    \centering
    \includegraphics[width=1\textwidth]{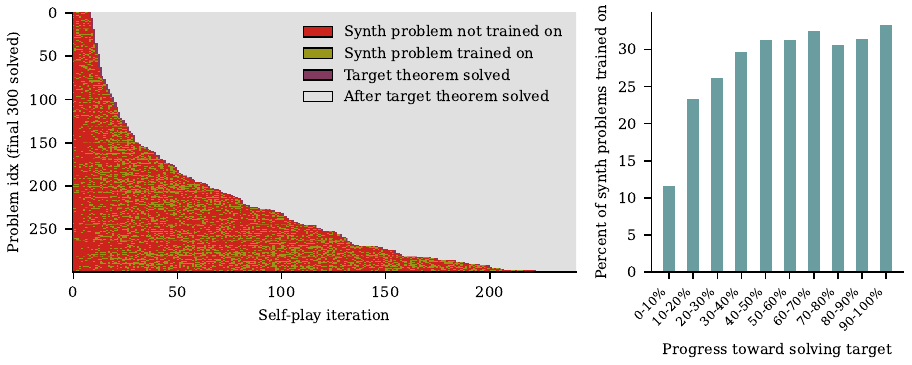}
    \vspace{-6mm}
    \caption{
        \textbf{More synthetic problems are solved as you approach solving a theorem.}
        \textbf{Left:} A heatmap of the last
        300 target theorems to be solved by \algoname in \Cref{fig:figure_1}.
        Each row shows the training history for a single target theorem. 
        \textbf{\textcolor{solved}{Green}} and \textbf{\textcolor{notsolved}{red}} cells
        mark iterations in which the theorem's synthetic problem is and is not
        suitable for training on respectively.
        \textbf{Right:} For each of the final 300 target theorems,
        we bin the pre-solve iterations into ten equally-sized buckets
        and compute the fraction of iterations in
        that bucket in which the theorem's synthetic problem is trained on. Bars
        show the average across the final 300 target theorems. We see as the \solver approaches solving each
        target problem, it is trained on more synthetic problems for said target.
    }
    \label{fig:conjecture_solved}
\end{figure}

\textbf{Main result.}
From \Cref{fig:baselines} we select \baseline as the best performing RL baseline.
We run \algoname on $D_{\text{3k}}$, and compare against 
this RL baseline and parallel sampling. The results are shown in \Cref{fig:figure_1}. 
Firstly, we see the performance of \algoname is consistently better than the 
RL baseline, with a 7\% higher asymptotic cumulative solve rate.
If we narrow our view to only the problems that 
the RL baseline never solves (of which there are 1346), we find \algoname
sees steady and sustained progress from the very beginning on such problems, 
solving almost 10\% after 8M generations compared to 0\% for RL after 4M generations  (see \Cref{fig:hard_run} of 
Appendix \Cref{sec:appendix_hard_problems}). 

As a yardstick to compare \algoname performance to, we use the pass@4 of the
much larger 671B parameter DeepSeek-Prover-V2 model.
 We see that at 6.3M generations,
\algoname applied to the 7B parameter DeepSeek-Prover-V2 model exceeds the pass@4 of the larger 671B 
counterpart. We also compare \algoname to STP \citep{dong2025stp}, the closest 
related self-play method designed \emph{specifically for Lean4 theorem
proving}. We find that \algoname has superior scaling, outperforming STP 
after 1M generations (\Cref{fig:stp}).

In \Cref{fig:main_run_reward} we show the training dynamics of \algoname. 
We see that over training, the \solver and \generator both improve reliably.
The \solver pass@8 on the target statements increases steadily, demonstrating
that the \solver is not suffering from catastrophic forgetting. For the \generator,
we see that the overall reward increases across training, primarily driven 
by a stable \reviewer reward, and increasing solve rate reward. 
In \Cref{fig:conjecture_solved} we show information about what 
synthetic problems are solved during training. Interestingly, 
from \Cref{fig:conjecture_solved} right panel, we see 
as the \solver approaches solving a target problem, it is trained 
on more synthetic problems for said problem. This is what one would 
expect should the synthetic problems for a specific target be 
useful for solving that target.

\begin{figure}
    \centering
    \includegraphics[width=1\textwidth]{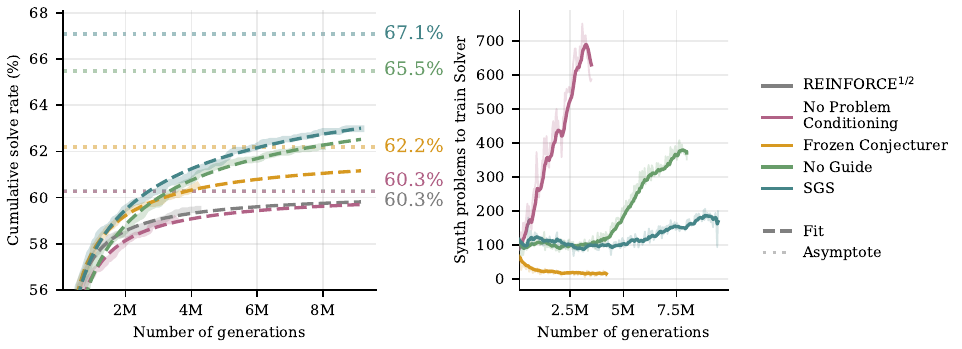}
    \caption{
        Ablation of the \algoname components.
        \textbf{Left:} Cumulative solve rate. The \reviewer improves
        performance at all points during training. Freezing the \generator
        leads to even worse performance, but still above baseline. 
        No Problem Conditioning shows no improvement over baseline.
        \textbf{Right:} Number of solvable problems used to
        train the \solver per iteration. Without the \reviewer, the
        \generator produces far more solvable problems, but this
        does not improve solve rate. With a frozen \generator, the
        fixed distribution of synthetic problems is quickly learned.
    }
    \label{fig:reviewer_study}
    \vspace{-1mm}
\end{figure}

\subsection{The benefit of self-guidance in the \generator algorithm}
\label{sec:reviewer_study}

We conduct a number of ablations on \algoname. Recall that 
\algoname has two components that influence the \generator 
distribution, (1) conditioning the \generator with unsolved 
problems, (2) using the \reviewer to reward the \generator. 
We ablate both and show the results in \Cref{fig:reviewer_study}.
All ablations use \baseline as the \solver objective and
are run on $D_{\text{3k}}$.\\

\begin{figure}[t]
    \centering
    \includegraphics[width=1\textwidth]{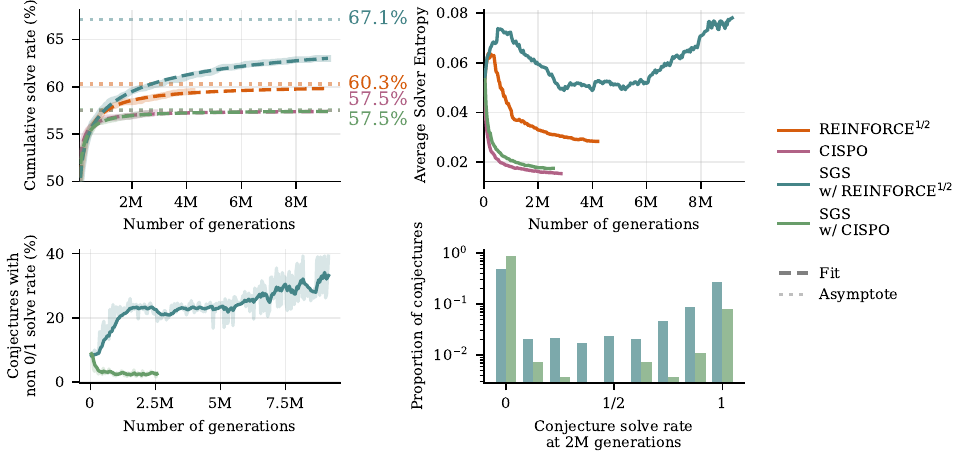}
    \caption{
        \algoname with CISPO vs.\ \baseline as the \solver objective.
        CISPO's entropy collapse (top right) concentrates problem solve
        rates at 0 and 1 (bottom right), starving the \generator of reward
        and preventing it from learning (bottom left). With
        \baseline, entropy remains stable and the \generator
        learns.
    }
    \label{fig:figure_entropy}
\end{figure}

\textbf{No Problem Conditioning.} This involves running 
the \generator without conditioning on unsolved problems,
and not using the \reviewer reward (we only reward 
the \generator using $R_{\mathrm{solve}}$).
We simply sample problems from the conjecturer with a 
fixed prompt (see Appendix \Cref{sec:appendix_prompting_details}),
however still sample 1 problem per unsolved target.
We see that this method 
\emph{does not outperform the RL baseline}. 
Indeed while the \generator is able to produce many problems 
that are difficult for the \solver (\Cref{fig:reviewer_study} right),
they are completely useless for solving more target problems. 

\textbf{No \reviewer.}  We compare the full \algoname algorithm against an identical
configuration in which we remove $R_{\text{guide}}$ from the \generator
reward, so that the \generator is trained only on
$R_{\text{solve}}$. We refer to this variant as ``No \reviewer.''
\algoname with the \reviewer achieves a higher cumulative solve rate than the
No \reviewer variant at every point during training, and the fitted asymptotic
solve rate is also higher: 67.1\% versus 65.5\%
(\Cref{fig:reviewer_study}, left). Both variants substantially outperform
the \baseline baseline and No Problem Conditioning.

Similar to No Problem Conditioning, the No \reviewer variant's lower performance
is surprising given that it
produces more training data for the \solver 
(\Cref{fig:reviewer_study},
right). Na\"ively, more solvable problems should mean more training signal
and faster learning. However, once again, \emph{the additional problems do not translate to 
better solve rates on target problems}.
We demonstrate why this is in \Cref{fig:qualitative}.
At  4M generations, when the No \reviewer \generator begins producing substantially more solvable
problems, it collapses to 
producing problems with long and complex conclusions, often 
involving many disjunctions (OR) of clauses. In \Cref{fig:qualitative} top, 
we show the number of disjunctive conclusions increasing and saturating 
to near 100\%, far higher than the base rate of less than 10\% present 
in the theorems of $D_{\text{3k}}$. In addition, \Cref{fig:qualitative} bottom 
shows the average length of conclusions increasing to nearly 10 times that 
of $D_{\text{3k}}$. In contrast to this, the \algoname \generator stably
produces problems with similar lengths and number of disjunctive conclusions as $D_{\text{3k}}$.
The \reviewer in \algoname ensures stability by down-weighting problems that are
superficially related to the target but inelegant.

\textbf{Frozen \generator.} Our prior ablation demonstrates that 
the \reviewer component helps to stabilize the long-running learning 
of the \generator, avoiding collapsing to problems that hack the
solve rate reward. Another much simpler method to stabilize the \generator
is to simply not train it. Indeed, this is the test time RL approach used
in AlphaProof \citep{hubert2025olympiad}, where a fixed Gemini model is used
as the \generator. We see that this method is better than the RL-baseline and 
No Problem Conditioning ablation,
but inferior to both the full \algoname method, and the No \reviewer ablation. 
Inspecting the \Cref{fig:reviewer_study} right panel, we can see why this may be the case. With a 
frozen \generator, the number of synthetic problems to train the \solver
decreases over time, demonstrating that the \solver capability quickly
covers the distribution of synthetic problems from the fixed \generator.
When the \solver is starved of synthetic problems, the algorithm reduces
to the RL baseline.

\subsection{Importance of managing entropy in the \solver algorithm}
\label{sec:entropy_collapse}

In \Cref{sec:algoname_performance} we observed that CISPO produces rapid entropy
collapse when used as an RL baseline. Here we show a general 
connection between the \solver{}'s entropy and \generator{}'s learning objective.

In \Cref{fig:figure_entropy} we show the training dynamics of \algoname
when using CISPO as the \solver objective. The asymptotic solve rate
remains \emph{essentially identical} to standalone CISPO. 
When CISPO is used as the \solver objective,
the \solver{}'s entropy collapses rapidly (\Cref{fig:figure_entropy}, top right). As the \solver becomes
near-deterministic, the distribution of problem solve rates concentrates
at 0 and 1 (\Cref{fig:figure_entropy}, bottom right).
This starves the \generator of any training signal, as problems
with 0 or 1 solve rate get 0 reward.
We observe different behavior when 
using \baseline as the \solver objective.
The \generator is able to learn to produce problems with a spread of solve rates
(\Cref{fig:figure_entropy}, bottom right). The presence
of synthetic problems from the \generator to train the \solver on
appears to stabilize the \solver{}'s entropy, which in turn
provides the \generator with an easier objective to learn.
Overall, we find that the entropy of the 
\solver remains stable when using \baseline.
We hypothesize that this connection between the \solver{}'s entropy
and \generator{}'s learning objective is key to making
successful LLM self-play algorithms. It does not preclude 
the use of grouped objectives like CISPO for the \solver, 
but additional methods to manage the entropy collapse
(such as the use of an entropy bonus or KL regularization to 
the base model) are required.

\section{Limitations and Conclusion}
\label{sec:discussion}

\textbf{Extension to domains beyond formal math}.
Our application of \algoname to the verifiable domain 
of formal math makes the role of the \generator easier
as it only has to produce an initial state (problem statements) 
as opposed to an entire MDP (including reward function).
We are excited about work extending \algoname to non-verifiable domains.
Consider, for example, applying \algoname to embodied control.
The \generator would need to specify a goal, a simulator or
environment in which the \solver can attempt that goal, and a reward
function. 
One could imagine the goal being generated by 
an LLM in text, environment being 
generated by a neural world model, like 
Google DeepMind's Genie \citep{bruce2024genie}, and a simple VLM
acting as the reward function. For coding problems,
the \generator could output a problem statement and unit tests. 
For natural language mathematics, learned verifiers have
improved rapidly \citep{luong2025towards} and could be 
used as-is to replace the formal verifier we use. 
More broadly, humans 
routinely learn from self-generated problems using only their
own (imperfect) ability to verify solutions, suggesting that 
perfect verification may not be strictly necessary for self-play
to be effective. 

\textbf{Learning the \reviewer}. In our experiments, the \reviewer is  
frozen.
While this is sufficient to prevent the \generator collapse we
observe in \cref{sec:reviewer_study}, it is likely insufficient for scaling 
\algoname to solve the most challenging problems. 
In such cases, the \reviewer must itself learn what constitutes 
a useful synthetic problem, since the characteristics of 
helpful stepping stones will evolve over self-play.
A natural extension is to train the \reviewer on 
aspects of the \solver training dynamics, for example labeling synthetic
problems that quickly lead to their corresponding target being 
solved as high reward.

\textbf{Scaling across model size}. Our experiments focus on scaling 
along the compute axis but hold model size fixed. This leaves 
open the question of how \algoname interacts with the model 
scale axis. We, naturally, hypothesize that the method 
will scale well in model size primarily because a larger 
\generator will produce higher-quality
synthetic problems. Confirming this hypothesis is an 
important direction 
for future work.

\subsection{Conclusion}

Asymmetric self-play is a promising paradigm for solving hard
problems far beyond a base model's capability. To solve the hardest 
problems, self-play must run for a long time, making it critical that learning scales
well with compute. Current methods scale poorly due to 
instability:
the \generator produces degenerate problems and the \solver
can undergo entropy collapse, starving the \generator of reward. \algoname
solves this: a \reviewer keeps synthetic problems grounded,
and the right \solver objective preserves \solver entropy.
The result is a self-play algorithm that sustains
learning for longer.

\section*{Acknowledgements}

LB thanks the support of a Stanford Graduate 
and Vitalik Buterin Fellowship.
TM thanks the support of NSF 2522743. KW thanks the support of the Stanford Graduate Fellowship. TH was supported by a grant by HAI, DSO labs, gifts from Open Philanthropy,
Amazon, Schmidt Sciences, the Tianqiao and Chrissy Chen Foundation and a grant under
the NSF CAREER IIS-2338866, ONR N00014-24-1-2609, and DARPA Cooperative Agreement HR00112520013. This work does not necessarily reflect the position or policy of the
government and no official endorsement should be inferred. We thank Google TPU Research Cloud 
and Stanford Marlowe \citep{kapfer2025marlowe} for computing resources. We
thank Thinking Machines for providing Tinker 
compute credits that were used in testing for various parts of this project.

We thank Lars Ankile and Tanishq Kumar 
for helpful feedback throughout the 
project.
We thank Tristan Thrush and Herman Brunborg for help testing our 
training infrastructure. We thank Neil Band, Caroline Choi, Thomas Chen, and Arvind Mahankali for feedback on an early draft of this work. 
We thank Mert Yuksekgonul and Yu Sun for useful conversations at the 
inception of the project, and writing feedback.
We thank Noah Jun and Marka Ellertson 
for help with manuscript writing.

\newpage
\bibliographystyle{colm2026_conference}
\bibliography{colm2026_conference}

\appendix
\newpage

\section{Fitting Cumulative Solve Rate}
\label{sec:appendix_fitting_cumulative_solve_rate}

Following other 
works modeling accuracy metrics \citep{khatri2025art, ruan2024observational}, we adopt a 
sigmoidal curve (with respect to log compute) of the form

\begin{align}
    R_C = R_0 + (A - R_0) \cdot \frac{1}{1 + (C_{\text{mid}} / C)^B},
\end{align}

where $R_C$ is the cumulative solve rate at
$C$ number of model generations (our proxy for compute),
$R_0$ is the initial solve rate, $A$ is
the asymptotic solve rate, $C_{\text{mid}}$ sets the midpoint of the curve,
and $B$ controls the rate at which performance approaches the asymptote.

We fit 
the parameters ${A, C_{\text{mid}}, B}$ by minimizing the sum of squared 
residuals using SciPy's \texttt{curve\_fit}. The initial performance $R_0$ is set 
to the first observed solve rate rather than treated as a free parameter.

Consistent with observations in both pre-training 
scaling analyses \citep{li2025mis} and RL scaling studies \citep{khatri2025art}
we find that the earliest data points, corresponding to the low-compute
regime, can disproportionately influence the fitted parameters. In our setting,
the first few iterations exhibit rapid gains that are not representative 
of the longer-term scaling trend. We therefore omit all data below 100,000
generations from the fit and re-center the remaining data, effectively modeling 
the cumulative gains after the initial low-compute phase. Empirically we 
found this increased the robustness of our fits.

\begin{figure}[ht]
    \centering
    \includegraphics[width=1\textwidth]{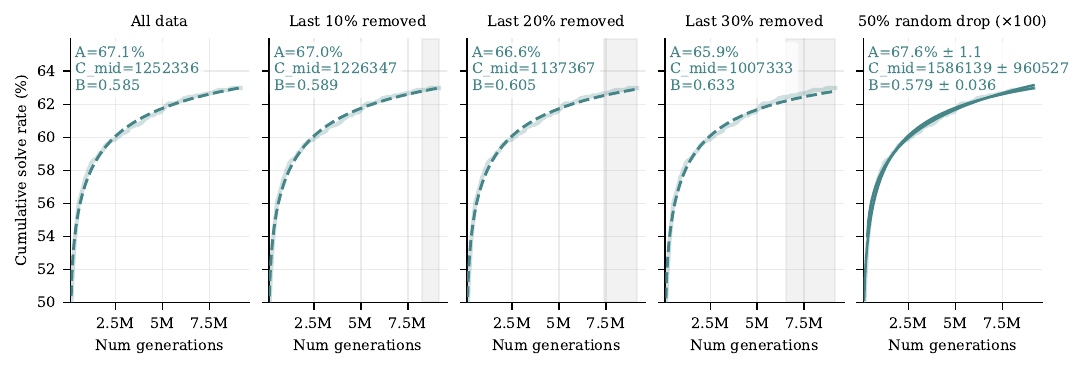}
    \caption{
        Sensitivity of the fitted scaling law to data removal for our main \algoname run.
        \textbf{Panels 1--4:} Fits with various amounts of training data removed 
        from the end of the run. The fitted values stay fairly robust, with a 1.6\% decrease 
        in asymptote value after 30\% of the ending data is removed.
        \textbf{Panel 5:} 100 independent fits, each using a uniformly sampled 50\% subset
        of the data (without replacement). The average and standard deviation 
        of fitted values is reported on the graph, as well as the fit with the highest 
        and lowest asymptotic value across all 100 fits. In this test, the fitted 
        values are also stable, with a 1.1\% standard deviation in the asymptotic value.
    }
    \label{fig:fit_sensitivity}
\end{figure}

\textbf{Validation of fits}. We validate the robustness of our fits in
two ways. First, we find that fits remain stable after removing 10\%, 20\%, and 30\% 
of the ending training data. Second we find the fits are robust to randomly dropping 
50\% of the training data, with the standard deviation of fitted asymptote values being 
only 1.1\% (see \Cref{fig:fit_sensitivity} for more details). This does 
however indicate that we should be cautious about drawing conclusions 
from asymptotic fits that are within 1.1\% of one another.
Of course, we cannot know the true asymptotic solve rate of the algorithms 
we test. We find the fitted scaling laws are a useful \emph{additional} 
metric to compare the performance of different algorithms, but not a 
substitute for real long-running performance, hence our focus on this 
kind of real data collection as well as curve fitting.

\newpage
\section{Training Infrastructure}
\label{sec:appendix_training_infrastructure}

Each iteration of \algoname alternates between a \emph{rollout stage}
(generation and verification) and a \emph{training stage}.
We coordinate two services during the rollout stage:
a GPU-backed generation service and a CPU-backed verification
service that checks proofs with the Lean4 compiler.

Both services follow a server--worker architecture: a central
HTTP server dispatches tasks in small batches to stateless workers
launched as independent Slurm jobs. When a worker requests a task, the
server assigns one from the pending queue. Completed results are
submitted back via HTTP POST. Generation and verification are fully
pipelined: completed proofs are forwarded for verification
immediately, so GPUs and CPUs stay saturated concurrently.

We implement two mechanisms to handle worker failures and stragglers.
First, automatic worker restart: when a worker reports itself
as dead (or times out), the server requeues its assigned tasks and
launches a replacement worker if work remains.
Second, speculative reassignment: when no pending tasks remain
but some are still in progress, the server assigns duplicate copies of
in-progress tasks to idle workers, preferring the task with the fewest
current workers. The first worker to return a result wins; the task is
removed from all other workers.

For generation, we use up to 64 GPU workers across multiple Slurm
partitions, each with 1 GPU and 32\,GB memory.
For verification, we use up to 128 CPU workers, each with 33 CPU
threads and 100 verification tasks per job.
Training uses a single H200 node with ZeRO Stage-2 distributed
optimization.
All models use bfloat16 precision and a maximum sequence length of
8192 tokens.

For verification, we use
Lean version 4.15.0, timeout set to 200s, and no memory limit.
We monitor the rate of verification system errors
throughout all runs, and ensure this never exceeds 1\% of
total verifications.

\begin{figure}[h]
    \centering
    \includegraphics[width=0.9\textwidth]{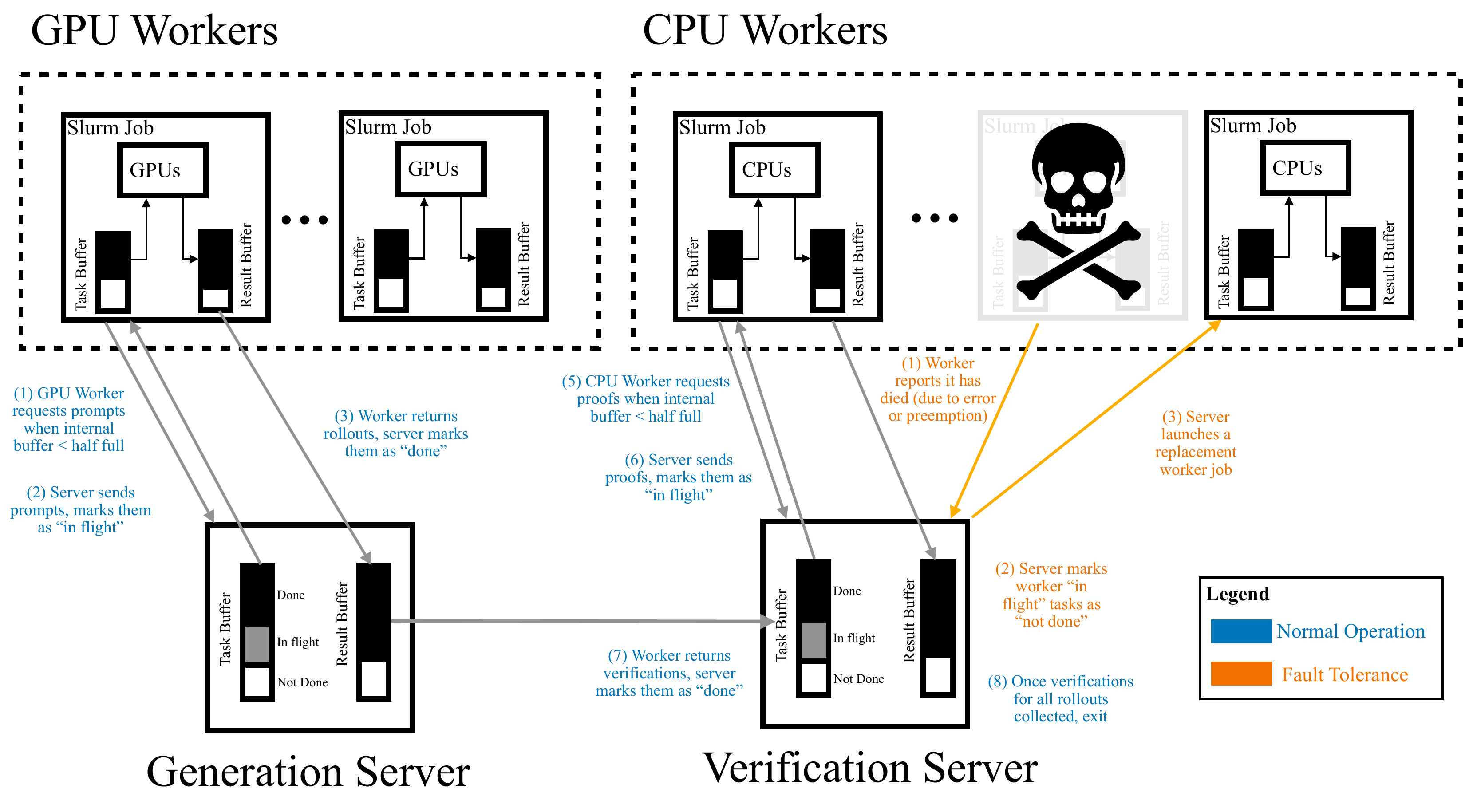}
    \caption{
        Diagram of inference infrastructure used for experiments. 
        We show the normal operation of the infrastructure (\textbf{\textcolor{infra-blue}{blue}}), 
        and the fault tolerance mechanisms (\textbf{\textcolor{infra-fault}{orange}}). 
        Note, the diagram shows the fault tolerance mechanisms for the
        Verification service, but it is used identically 
        for the Generation service.
    }
    \label{fig:mp}
\end{figure}

\newpage
\section{Hyperparameters}

Here we include details of the hyperparameters used 
in our experiments, and the selection mechanisms.

\subsection{Shared Hyperparameters}
\label{sec:appendix_shared_hyperparameters}

All methods use the Adam optimizer with $\beta_1 = 0.9$,
$\beta_2 = 0.95$, gradient clipping at norm 1.0, and a
constant learning rate schedule.
We use a batch size of 32
(per-GPU batch size of 1, with ZeRO Stage-2 data parallelism).
The maximum sequence length is 8192 tokens.
All generation uses a temperature of 1.0.
For CISPO, we set $\epsilon_{\mathrm{low}} = 1.0$ and
$\epsilon_{\mathrm{high}} = 3.0$ (clipping the importance weight
to $[0, 4]$).

\subsection{Learning Rate Selection}

To ensure the strongest possible baselines, we sweep over
learning rates $\{3\text{e-}7,\; 1\text{e-}6,\;
3\text{e-}6,\; 1\text{e-}5\}$ for all three
RL baselines on $D_{\text{3k}}$, selecting the learning rate
with the highest cumulative solve rate after 500k generations.
In all cases we found the best performing learning rate was 
not boundary point of these values (we originally did 
not test 1e-5, however expanded to include this value 
when 3e-6 appeared to be optimal for \baseline).
The optimal learning rate for CISPO is $1\text{e-}6$,
while both EI and \baseline perform best at $3\text{e-}6$.
For \algoname, we inherit the $3\text{e-}6$ learning rate
from the \baseline sweep, using it for both the \solver and
\generator.

\subsection{Individual Experiment Hyperparameters}

All experiments use the shared hyperparameters from the preceding
subsection and the learning rates described in the learning rate
selection subsection. We provide a summary
of per-experiment configurations in \Cref{tab:experiment_hypers}.

\begin{table}[h]
\centering
\small
\begin{tabular}{lccccc}
\toprule
\textbf{Experiment} & \textbf{Figure(s)} & \textbf{LR} & \textbf{\solver Obj.} & \textbf{Synth.} & \textbf{\reviewer} \\
\midrule
\algoname & \ref{fig:figure_1}, \ref{fig:reviewer_study}, \ref{fig:figure_entropy} & $3\text{e-}6$ & \baseline & 1 & \checkmark \\
\baseline & \ref{fig:figure_1}, \ref{fig:baselines}, \ref{fig:reviewer_study}, \ref{fig:figure_entropy} & $3\text{e-}6$ & \baseline & 0 & --- \\
Expert Iteration & \ref{fig:baselines} & $3\text{e-}6$ & EI & 0 & --- \\
CISPO & \ref{fig:baselines}, \ref{fig:figure_entropy} & $1\text{e-}6$ & CISPO & 0 & --- \\
\midrule
No \reviewer & \ref{fig:reviewer_study} & $3\text{e-}6$ & \baseline & 1 & \ding{55} \\
Frozen \generator & \ref{fig:reviewer_study} & $3\text{e-}6$ & \baseline & 1 & \ding{55} \\
No Problem Cond. & \ref{fig:reviewer_study} & $3\text{e-}6$ & \baseline & 1 & \ding{55} \\
\midrule
CISPO \solver & \ref{fig:figure_entropy} & $1\text{e-}6$ & CISPO & 1 & \checkmark \\
\bottomrule
\end{tabular}
\caption{
    Per-experiment hyperparameters.
    \textbf{Synth.}: number of synthetic problems generated per
    unsolved target statement.
    All other hyperparameters (optimizer, batch size, gradient clipping,
    sequence length, CISPO clipping bounds) are as described
    in the shared hyperparameters subsection.
}
\label{tab:experiment_hypers}
\end{table}

\newpage
\section{Data Details}
\label{sec:appendix_data_details}

We detail the method we use to create the $D_{\text{3k}}$ dataset. 
We begin with Goedel-Pset-V1 \citep{lin2025goedel1}, a 
large collection of auto formalized 
problems primarily from the NuminaMath-CoT 
dataset \citep{li2024numinamath}. We conduct the following 
filtering steps:

\begin{enumerate}
    \item \textbf{First pass removing easy problems.} Remove any problem with a pass rate of 
    above 50\% (estimated from 8 rollouts per proof)
    using the STP 7B lean prover model from 
    \citet{dong2025stp}. 
    \item \textbf{First pass removing false problems, second 
    pass removing easy problems.}
    Next we take a slightly more capable prover, Kimina-Prover-Preview
    (trained with a large-scale reinforcement learning pipeline from Qwen2.5-72B
    \citep{wang2025kimina})
    and do 8 rollouts for every problem and for
    \emph{the negation of every problem}. If it is possible to prove
    the negation of the problem, then the original statement was 
    false (and thus unprovable). We remove any problem 
    where the negation was proved at least once, and as a second 
    pass removing easy problems, we remove any problems that 
    Kimina-Prover-Preview proves more than 2/8 times.
    \item \textbf{Subsampling.} As we are interested in 
    studying a regime with many rounds of self-play, we 
    subsample the dataset down to 5,000 problems randomly.
    \item \textbf{Final pass removing false problems.}
    Finally, we take the remaining 5,000 problems, and
    prompt GPT 5 mini to
    identify any problems that it thinks are false.
    We prompt the model to reason and then output
    one of `solvable', `unknown', `unsolvable'. We remove
    any problems given the label `unsolvable'. This
    leaves us with 3,323 problems, which make
    up $D_{\text{3k}}$. The full prompt is provided in
    \Cref{sec:appendix_solvability_filtering_prompt}.
\end{enumerate}

\begin{figure}[h]
    \centering
    \includegraphics[width=1\textwidth]{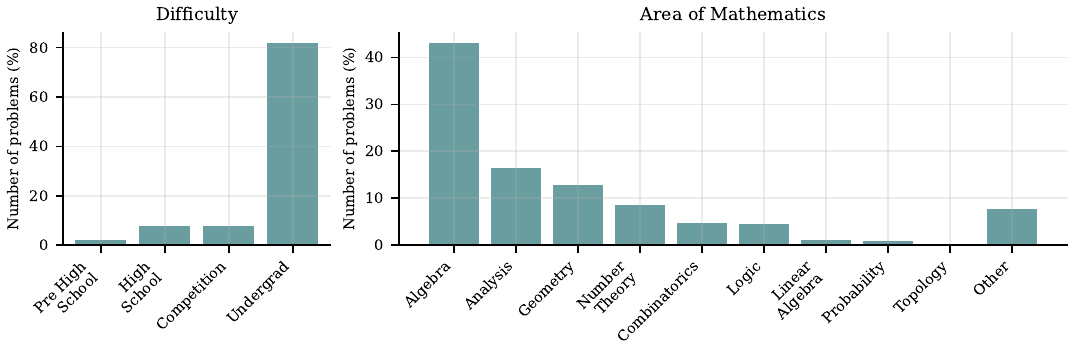}
    \caption{
        Distribution of problem 
        types and difficulty 
        in $D_{\text{3k}}$.
    }
    \label{fig:d_3k}
\end{figure}

\subsection{Composition of $D_{\text{3k}}$}

We release $D_{\text{3k}}$ publicly. \footnote{$D_{\text{3k}}$ can 
be found at \url{https://huggingface.co/datasets/LukeBailey181Pub/D_3k}}
We provide details about the dataset composition in
\Cref{fig:d_3k}. Specifically, for each
problem in $D_{\text{3k}}$ we prompt GPT 5.4 nano
to categorize the difficulty of the problem
as one of \{Pre High School, High School, Competition, Undergraduate\}
and topic as one of
\{Algebra, Analysis, Geometry, Number Theory, Combinatorics,
Logic, Linear Algebra, Probability, Topology, Other\}.
For any problem that GPT 5.4 nano
provides an output that we cannot parse, we reprompt
until we have categorizations for over 99\% of the dataset.
We provide details of the prompt used for this
categorization in \Cref{sec:appendix_dataset_categorization_prompt}.

\newpage
\section{Prompting Details}
\label{sec:appendix_prompting_details}

We describe the prompts used for the \generator and \reviewer roles
in \algoname when using DeepSeek-Prover-V2-7B as the base model.

\subsection{\generator Prompt}

The \generator is prompted with an unsolved target problem and asked to
produce a related lemma or theorem. The full prompt is as follows:

\begin{quote}
\ttfamily\small
Here is a Lean 4 problem statement:\\

\textasciigrave\textasciigrave\textasciigrave lean4\\
\{target\_theorem\}\\
\textasciigrave\textasciigrave\textasciigrave\\

Please generate a lean4 theorem that is a lemma or related theorem
that is useful for proving the above statement. It should be possible
to use the lemma or related theorem to help prove the above statement.
The lemma or related theorem should be simpler to prove than the
target statement. It should NOT be identical to the target statement.
It should NOT be equivalent through renaming variables and premises.\\

Output the final theorem as a syntactically correct lean4 theorem
statement between \textasciigrave\textasciigrave\textasciigrave lean4 and \textasciigrave\textasciigrave\textasciigrave\ tags.
The final thing you output should be the theorem statement WITHOUT
any proof, just put `sorry' for the proof.
\end{quote}

\subsection{No Problem Conditioning Prompt}

In the No Problem Conditioning ablation (\Cref{sec:reviewer_study}), the \generator is not
conditioned on any target problem. Instead, it is prompted to generate
an arbitrary theorem:

\begin{quote}
\ttfamily\small
Please generate a Lean4 theorem statement. The theorem should be
non-trivial, but not too difficult to prove. Choose a random area of
mathematics for the theorem, such as algebra, number theory,
combinatorics, geometry, calculus, or logic. The theorem should be
interesting and mathematically meaningful, not just a very easy identity
or direct restatement of a definition.\\

Output the final theorem as a syntactically correct Lean4 theorem
statement between \textasciigrave\textasciigrave\textasciigrave lean4 and \textasciigrave\textasciigrave\textasciigrave\ tags.
The final thing you output should be the theorem statement WITHOUT
any proof, and just put `sorry' for the proof.
\end{quote}

\subsection{\reviewer Prompt}

The \reviewer is a finetuned copy of DeepSeek-Prover-V2-7B that scores
each synthetic problem on three dimensions. We finetune on 2048 examples
of well-formatted outputs generated by GPT-4.1 mini to ensure reliable
output formatting (see \Cref{sec:implementation}).
Given a target theorem and a generated conjecture, the \reviewer is
prompted as follows:

\begin{quote}
\ttfamily\small
You are a math expert. Here is a seed lean4 problem statement:\\

\textasciigrave\textasciigrave\textasciigrave lean4\\
\{seed\_theorem\}\\
\textasciigrave\textasciigrave\textasciigrave\\

Here is a lemma or related lemma that is supposed to be useful for proving the above statement.
It can be useful in that either it is a lemma that can be directly used to help prove the above
statement, or it is a related lemma that plausibly requires similar proof techniques to solve.\\

\textasciigrave\textasciigrave\textasciigrave lean4\\
\{conjecture\}\\
\textasciigrave\textasciigrave\textasciigrave\\

Please rate the relevance of the lemma to the seed problem on a scale of 0 to 5, where 0 is
``not at all related'' and 5 is ``very useful for proving the target statement.'' If the lemma
is trivial, you should give it a low score.\\

Here is a rubric for how to score the lemma:\\
- 0: The lemma is not at all related to the seed problem and is trivial to prove.
OR the lemma is identical (including equivalent by renaming of premises and variables)
to the seed problem.\\
- 1: The lemma is not at all related to the seed problem.\\
- 2: The lemma is related to the seed problem in that it concerns a similar subfield
of mathematics, but is not directly useful for proving the seed problem.\\
- 3: The lemma is related to the seed problem and may be useful for proving the seed problem.\\
- 4: The lemma is directly useful for solving the seed problem. That is if the lemma was proved,
the seed problem would be easier to solve.\\
- 5: The lemma is very useful for solving the seed problem, and solving the lemma will
dramatically reduce the difficulty of the original seed problem.\\

Next decide how redundant the premises are. Rate this as 0 or 1:\\
- 0: There are no redundant premises.\\
- 1: There are redundant premises. That is premises that are not needed to prove the conclusion.\\

Next decide if the conclusion is overly complex. Rate this on a score of 0 to 4:\\
- 0: The conclusion is minimally complex. A single, atomic statement that is maximally clear
and easy to apply to other problems.\\
- 1: The conclusion has low complexity. Multiple related parts (e.g., 2--3 conjunctions) but
they form a cohesive statement.\\
- 2: The conclusion has moderate complexity. Contains disjunctions but they are closely related
alternatives, or contains multiple conjunctions (3--4) that are all on-theme.\\
- 3: The conclusion has high complexity. Contains multiple unrelated or clauses (2--3 disjunctions),
or contains deep nesting of logical operators that obscure the main claim.\\
- 4: The conclusion has very high complexity. A disjunction of many (3 or more) largely unrelated
clauses, or contains deeply nested logical structure that is hard to parse.
\end{quote}

\noindent
The three sub-scores are combined into a single review score as follows.
If the conclusion complexity is 3 or 4 (high or very high), the review
score is automatically 0. Otherwise:
\begin{align}
    R_{\text{guide}} = \max\Big(0,\; \text{relevance} + (2 - \text{complexity}) + (1 - \text{redundancy})\Big).
\end{align}
This rewards relevant conjectures with simple, non-redundant conclusions,
and assigns zero reward to conjectures with overly complex conclusions.

\subsection{Solvability Filtering Prompt}
\label{sec:appendix_solvability_filtering_prompt}

During dataset construction (\Cref{sec:appendix_data_details}), we use
GPT 5 mini to filter out unsolvable problems. Given a candidate problem,
the model is prompted as follows:

\begin{quote}
\ttfamily\small
Here is a lean4 problem:\\

\textasciigrave\textasciigrave\textasciigrave lean4\\
\{theorem\}\\
\textasciigrave\textasciigrave\textasciigrave\\

I want you to determine if this problem is solvable. There are three possible answers:\\

1) Yes, the problem is solvable. That is there exists some proof of the problem. Note that if two premises contradict each other, then the problem IS solvable as it is vacuously true.\\
2) Unknown. You are not sure if the problem is solvable or not.\\
3) No, the problem is not solvable. I.e.\ the conclusion is not provable from the premises.\\

Reason about the problem and then decide on an answer from the above. If your answer is 1), output <SOLVABLE>.
If your answer is 2), output <UNKNOWN>. If your answer is 3), output <UNSOLVABLE>.\\

Make sure to only output one of these three tags.
\end{quote}

\noindent
Problems receiving the \texttt{<UNSOLVABLE>} label are removed from the dataset.

\subsection{Dataset Categorization Prompt}
\label{sec:appendix_dataset_categorization_prompt}

To categorize the difficulty and topic of each problem in $D_{\text{3k}}$
(\Cref{sec:appendix_data_details}), we use GPT 5.4 nano with the
following prompt:

\begin{quote}
\ttfamily\small
Classify this Lean 4 theorem. Respond with ONLY two words: difficulty area. No reasoning.\\

Difficulty: pre\_high\_school, high\_school, competition, undergraduate\\
Area: algebra, analysis, number\_theory, combinatorics, geometry, topology, probability, linear\_algebra, logic, other\\

Example response: undergraduate algebra\\

Theorem:\\
\{theorem\}
\end{quote}

Any problems we cannot parse the answer from, we
reprompt until we have over 99\% of the dataset classified.

\newpage
\section{Additional Results}
\label{sec:appendix_additional_results}

In this section we provide additional results 
from the experiments conducted in \Cref{sec:ex}.

\subsection{Baseline Entropy}

\begin{figure}[h]
    \centering
    \includegraphics[width=0.5\textwidth]{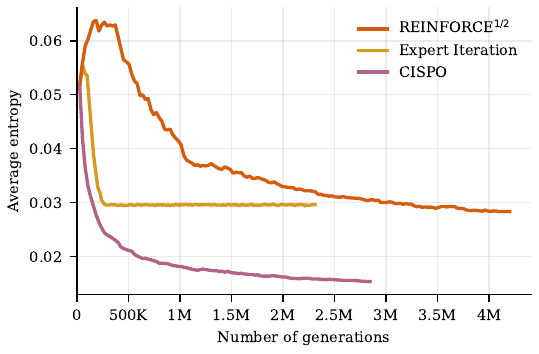}
    \caption{
        Mean per-token entropy of the \solver{}'s output distribution for each
        RL baseline on $D_{\text{3k}}$. CISPO collapses fastest, followed
        by Expert Iteration which plateaus at a low entropy floor early in
        training. Note the entropy values are naturally lower than one would expect 
        from LLMs as most of the rollouts are Lean specific tokens.
    }
    \label{fig:baselines_entropy}
\end{figure}

\Cref{fig:baselines_entropy} is the companion 
figure to \Cref{fig:baselines}, showing the entropy of the \solver{}'s output
 distribution for each baseline method. We see that 
 CISPO suffers the fastest entropy collapse, 
 \baseline entropy also decreases but at a slower rate, 
 and EI entropy plateaus as late in training the number 
 of problems it trains on becomes extremely small.

\subsection{Comparison to STP}
\begin{figure}[h]
    \centering
    \includegraphics[width=0.5\textwidth]{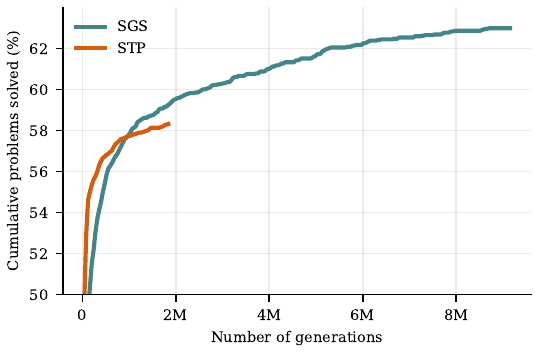}
    \caption{
        \algoname and STP \citep{dong2025stp} performance on 
        $D_{\text{3k}}$. We see that \algoname has superior 
        scaling, with a crossover happening at around 1M generations.
    }
    \label{fig:stp}
\end{figure}

We compare \algoname to STP \citep{dong2025stp}, the closest 
related self-play method designed \emph{specifically for Lean4 theorem
proving}. To run STP we take the code released by \citep{dong2025stp}. 
This code is designed to run on TPUs, so we use 64 v4 TPUs 
as opposed to GPUs. To get better hardware utilization, we 
also increase the batch size to 64. For hyperparameter
optimization, we go through a similar process to \algoname. In particular,
the Solver algorithm used in STP is EI, and thus we take the optimal 
EI learning rate we found of $3\text{e-}6$, and keep the 
rest of the hyperparameters the same as described in \Cref{sec:appendix_shared_hyperparameters}.
The results are shown in \Cref{fig:stp}. We see 
that \algoname has superior scaling, with a crossover 
happening at around 1M generations.

\subsection{Performance on problems beyond RL Baseline}
\label{sec:appendix_hard_problems}

\begin{figure}[h]
    \centering
    \includegraphics[width=0.5\textwidth]{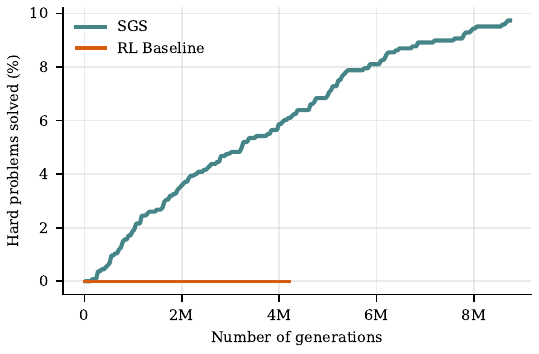}
    \caption{
    Performance on $D_{\mathrm{hard}}$, the subset of $D_{\mathrm{3k}}$ never solved by \baseline, during the main run in \Cref{fig:figure_1}. \algoname steadily solves a growing fraction of these problems.
    }
    \label{fig:hard_run}
\end{figure}

Taking the RL baseline (\baseline) and \algoname run data from 
\cref{fig:figure_1}, we derive a new dataset $D_{\mathrm{hard}} \subset D_{\mathrm{3k}}$ 
of all the problems that the RL baseline never solved. We get 
$|D_{\mathrm{hard}}| = 1346$, roughly 1/3 the size of $D_{\text{3k}}$. 
In \Cref{fig:hard_run} we show the percent of problems in 
$D_{\mathrm{hard}}$ that are solved by \algoname and the RL baseline 
during the main experimental run (\cref{fig:figure_1}). By construction,
the RL baseline solves 0 problems. However, \algoname shows steady 
progress on this set of hard problems that are inaccessible to RL,
reaching close to a 10\% cumulative solve rate. 

 $D_{\mathrm{hard}}$ isolates the difficult problems 
that (within 4M generations) our best RL method gets \emph{0 reward on}. 
Despite this, we see \algoname making steady and non-trivial progress on such problems
from the very beginning of the run.

\newpage
\section{RL Objective Functions}
\label{sec:appendix_rl_objective_functions}

We describe the RL objectives used in our experiments.
At each iteration, we sample a batch of problems
$\mathcal{B} \subseteq \mathcal{D}$ and generate $k$ rollouts
$\{y^i\}_{i=1}^k$ per problem $x \in \mathcal{B}$.
Each rollout is verified by the Lean4 compiler, yielding a binary
reward $v(y) \in \{0,1\}$.

\subsection{REINFORCE}

The base REINFORCE objective  is a
log-likelihood update weighted by the binary reward:

\begin{equation}
\begin{aligned}
\mathcal{L}_{\mathrm{RF}}(\theta)
=&
- \mathbb{E}_{x \sim \mathcal{B},\; y \sim \pi_{\theta_{\mathrm{old}}}(\cdot \mid x)}
\\
&\Biggl[
\frac{1}{|y|}
\sum_{t=1}^{|y|}
v(y)\,
\log \pi_\theta \bigl( y_t \mid x, y_{<t} \bigr)
\Biggr]
\end{aligned}
\end{equation}

\subsection{\baseline}

\baseline is our primary \solver objective. It applies the
REINFORCE loss above, but only to problems with solve rate
$s(x) = \frac{1}{k}\sum_{i=1}^k v(y_x^i) \leq 0.5$.
All problems with $s(x) > 0.5$ are discarded from the
training batch. This prevents the \solver from spending
gradient updates on already-easy problems, focusing
learning on problems where progress is still needed.

\subsection{CISPO}

CISPO is a grouped reinforcement learning objective introduced
by \citet{chen2025minimax}. The recent study by \citet{khatri2025art}
found it to have the best asymptotic performance of LLM RL algorithms
developed in the past year. Like GRPO \citep{shao2024deepseekmath},
CISPO uses the group of rollouts $\{y^i\}_{i=1}^k$ from a single
prompt $x$ to calculate an advantage, however it uses an improved
importance sampling clipping method compared to the widely used
PPO-style clipping \citep{schulman2017proximal}.

For a problem $x$ and its rollout group $\{y^i\}_{i=1}^k$,
where each rollout $y^i = (y_{i,1}, \dots, y_{i,|y^i|})$,
let
\[
w_{i,t}
=
\frac{
\pi_\theta \bigl( y_{i,t} \mid x, y_{i,<t} \bigr)
}{
\pi_{\theta_{\mathrm{old}}} \bigl( y_{i,t} \mid x, y_{i,<t} \bigr)
}
\]
be the token-level importance sampling weight. Let
\begin{align*}
&\hat{A}^i
=
\frac{
v(y^i) - \mathrm{mean}(\{v(y^j)\}_{j=1}^k)
}{
\mathrm{std}(\{v(y^j)\}_{j=1}^k)
}
\quad \text{and} \\
&\hat{w}_{i,t}
=
\mathrm{clip}
\bigl(
w_{i,t},\;
1 - \epsilon_{\mathrm{low}},\;
1 + \epsilon_{\mathrm{high}}
\bigr)
\end{align*}
be the group-relative advantage and clipped token-level importance
sampling weight respectively, where $\epsilon_{\mathrm{low}}$ and
$\epsilon_{\mathrm{high}}$ are hyperparameters. The CISPO loss is:

\begin{equation}
\begin{aligned}
&\mathcal{L}_{\mathrm{CISPO}}(\theta)
=- \mathbb{E}_{x \sim \mathcal{B},\; \{y^i\}_{i=1}^k \sim \pi_{\theta_{\mathrm{old}}}(\cdot \mid x)}
\\
&\Biggl[
\frac{1}{\sum_{i=1}^k |y^i|}
\sum_{i=1}^k
\sum_{t=1}^{|y^i|}
\operatorname{sg}\!\left( \hat{w}_{i,t} \right)
\hat{A}^i\,
\log \pi_\theta \bigl( y_{i,t} \mid x, y_{i,<t} \bigr)
\Biggr]
\end{aligned}
\end{equation}

\noindent
where $\mathrm{sg}$ is the stop-gradient operator.

\subsection{Expert Iteration}

Our Expert Iteration (EI) baseline, following \citet{dong2025stp},
only generates rollouts for problems that have been solved fewer than
16 times in prior iterations. Training is performed on correct proofs
from the last 3 iterations using the REINFORCE objective.
This focuses compute on problems where
the model has limited coverage, but we find that late in training
the number of newly solved problems per iteration tends to zero,
causing learning to stall.

\subsection{\generator Objective}
\label{sec:app_conj_objective}

The \generator is trained with REINFORCE using the combined reward
$R_{\text{synth}} = R_{\text{solve}} \cdot R_{\text{guide}}$, as
described in \Cref{sec:method}. The rewards are linearly normalized
to $[0, 1]$ within each batch before computing the policy gradient.
We do so  
by replacing $r_i$ with $(r_i - \min_j r_j )/(\max_j r_j - \min_j r_j)$.
Conjectures with a solve rate of 0 (unsolvable) or in the top 30\%
of solve rates within the batch (too easy) receive
$R_{\text{solve}} = 0$.

\end{document}